\documentclass[11pt]{article}

\usepackage[preprint]{acl}

\usepackage{times}
\usepackage{latexsym}
\usepackage[T1]{fontenc}
\usepackage[utf8]{inputenc}
\usepackage{microtype}
\usepackage{booktabs}
\usepackage{tabularx}
\usepackage{enumitem}
\usepackage[table]{xcolor}
\usepackage{tikz}
\usetikzlibrary{positioning,fit}
\usepackage[capitalise,noabbrev]{cleveref}

\makeatletter
\def\ACL@lnx#1#2{\expandafter\gdef\csname ACL@lnx@#1\endcsname{#2}}
\@ifundefined{linenumbersep}{}{%
  \setlength{\linenumbersep}{6pt}%
  \setlength{\linenumberwidth}{18pt}
  \let\ACL@LNleft\makeLineNumberLeft
  \let\ACL@LNright\makeLineNumberRight
  \def\makeLineNumber{%
    \pdfsavepos
    \edef\ACL@tmp{\write\@auxout{\string\ACL@lnx{\the\c@linenumber}{\noexpand\the\pdflastxpos}}}%
    \ACL@tmp
    \@ifundefined{ACL@lnx@\the\c@linenumber}%
      {\if@firstcolumn\ACL@LNleft\else\ACL@LNright\fi}%
      {\ifdim\csname ACL@lnx@\the\c@linenumber\endcsname sp<.5\paperwidth\relax
         \ACL@LNleft\else\ACL@LNright\fi}}%
  \AtBeginDocument{
                   
                   }%
}
\makeatother

\definecolor{substrate}{HTML}{56B4E9}
\definecolor{capability}{HTML}{009E73}
\definecolor{evidence}{HTML}{E69F00}
\definecolor{rowtint}{HTML}{EFF3F7}

\hypersetup{pdftitle={Evaluating Real-Time Voice Agents: A Survey from Component Quality to Grounded Outcomes},
  pdfauthor={Shivam Negi, Arpit Rawat, Rashi Jain}}

\newcommand{\NArtifacts}{140}
\newcommand{\NPrimary}{136}
\newcommand{\NPositioning}{3}
\newcommand{\NMethod}{1}
\newcommand{\NPrimaryRecent}{90}
\newcommand{\NPrimaryArxiv}{118}
\newcommand{\NPrimaryPublished}{41}
\newcommand{\NEvalPublished}{31}
\newcommand{\NStatePublished}{3}
\newcommand{\NCapability}{136}
\newcommand{\NEval}{91}

\newcommand{\NEvalMeta}{102}
\newcommand{\NState}{13}
\newcommand{\NStateSim}{10}
\newcommand{\NStateRecent}{11}
\newcommand{\NGrounded}{21}
\newcommand{\NGroundedRS}{5}
\newcommand{\NMulti}{26}
\newcommand{\NMultiState}{2}

\newcommand{\NRecovery}{35}
\newcommand{\NRecoveryAV}{10}

\newcommand{\NPool}{2472}
\newcommand{\NExpIncluded}{49}
\newcommand{\NExpExcluded}{2}
\newcommand{\NExpRetained}{47}

\newcommand{\NSensPeerEval}{31}
\newcommand{\NSensPeerState}{3}
\newcommand{\NSensPeerGrounded}{3}
\newcommand{\NSensPeerGroundedRS}{2}

\newcommand{\NSensPreEval}{31}
\newcommand{\NSensPreState}{2}
\newcommand{\NSensPreGrounded}{4}
\newcommand{\NSensPreGroundedRS}{3}

\newcommand{\NSensBaseEval}{36}
\newcommand{\NSensBaseState}{7}

\makeatletter


\makeatother

\title{Evaluating Real-Time Voice Agents: A Survey from Component Quality to Grounded Outcomes}

\author{Shivam Negi\textsuperscript{1} \quad Arpit Rawat\textsuperscript{1,2} \quad Rashi Jain\textsuperscript{3}\\
  \textsuperscript{1}Northeastern University, Boston, MA, USA \quad
  \textsuperscript{2}Sinch Mailgun\\
  \textsuperscript{3}University of Texas at Dallas, Richardson, TX, USA\\
  \texttt{negi.sh@northeastern.edu}, \texttt{rawat.a@northeastern.edu}, \texttt{rxj210013@utdallas.edu}}

\begin{document}
\maketitle

\begin{abstract}
Real-time voice agents must decide when to speak, recover from disruption, and
act only when the conversation requires it. But currently their evaluation is
fragmented and is split across speech quality, turn-taking, tool calling, and
task completion. We review \NPrimary{} primary works published up to October 2,
2026, and code each on two axes: the capability an evaluation exercises, namely
timing, recovery, grounded outcome, and, where relevant, multiparty conduct;
and the strength of its evidence, namely evaluation conditions and grading
instrument. The coding locates the gaps. Only \NState{} of \NEval{} evaluation
works check external state or action traces, only \NGroundedRS{} of
\NGrounded{} grounded-outcome benchmarks use real human speech, and only
\NMultiState{} of \NMulti{} multiparty benchmarks verify state. Three claims
follow. Architecture constrains deployment but does not determine results. It
has been observed that benchmarks can verify outcomes, but few do. Evaluating
an agent solely through one-on-one user interaction leaves critical gaps: it
cannot confirm if the agent is speaking to the correct person, verify that
person's authority, or identify the proper moments for strategic silence.

From the taxonomy we derive TRG, short for Timing-Recovery-Grounded, a minimum
reporting standard, and apply it to three published reports with different
architectures. In one, tool-selection F1 is 82.5\% while exact-call success is
33.0\%, with no evidence of external task state. None of the three reports
measures task state after an interruption. TRG governs what is reported and
assigns no composite score. Per-source codes and tooling accompany the
submission.
\end{abstract}

\section{Introduction}
Spoken interaction poses an evaluation problem that ordinary question answering
does not. An agent can understand the words, choose the correct tool, and still
fail because a bystander spoke the command. It can respond quickly and still
interrupt a hesitation; it can remain politely silent when intervention is
needed. Each failure concerns whether a response or action fits its
conversational context, which answer quality alone cannot settle. VGBench,
TACT, and necessity-sensitive floor-taking evaluation expose these distinctions
directly \citep{zhang2026vgbench,shamsaie2026tact,floor2026necessity}.

The pipeline of automatic speech recognition (ASR), language modeling, and
text-to-speech (TTS) makes intermediate decisions inspectable, but
conventional silence-based segmentation restricts overlap. End-to-end speech
models instead represent simultaneous listening and speaking
\citep{defossez2024moshi,veluri2024synchronous}. Architecture alone does not
determine interaction quality: DuplexCascade supplies duplex control within a
cascade \citep{duplexcascade2026}, and hybrid delegation separates the
conversational frontend from tool execution \citep{hu2026frontendbackend}.
The open question is which evidence warrants a claim that such a system
behaves well. We make three contributions.
\begin{enumerate}[leftmargin=1.3em,itemsep=1pt,topsep=2pt]
\item A two-axis taxonomy that separates the capability an evaluation exercises
from the conditions and instrument that qualify its evidence. All \NArtifacts{} source
artifacts receive a review role, the \NCapability{} primary works that exercise a
capability receive Axis~A codes, and the \NEvalMeta{} evaluation and meta-evaluation
works receive the full Axis~B coding (\cref{sec:method,app:coding}).
\item A coverage analysis and a benchmark map, both generated from the coding,
that locate where the evidence is thin (\cref{tab:coverage,app:coding}).
\item TRG, a minimum reporting contract derived from the taxonomy, together with
worked audits of three published reports that show which gaps it exposes
(\cref{ssec:trg}).
\end{enumerate}
The contribution is methodological: we introduce neither a benchmark nor a
pooled performance ranking, and we do not re-run the reviewed systems.

\paragraph{Relation to existing surveys}
SpeechLM construction \citep{cui2024speechlm} and direct speech translation
\citep{gupta2024s2st} supply background with different objects of evaluation.
The closest contemporary work is \citet{lu2026duplexsurvey}. Its authors report
that it was accepted to EMNLP 2026 as \emph{Speaking While Listening: A Survey
and Empirical Audit of Full-Duplex Spoken Dialogue Systems}; we analyzed the
arXiv version, which still carries the earlier title. \Cref{tab:lu} contrasts
the two. Lu et al.\ ask which states a system reaches; we ask which deployment
claims a reported result supports. The two are complementary: their
interaction ontology can specify the test conditions, and TRG can specify what
a report of those tests must contain.

\begin{table}[t]
\centering
\footnotesize
\setlength{\tabcolsep}{3pt}
\begin{tabularx}{\columnwidth}{@{}l>{\raggedright\arraybackslash}X>{\raggedright\arraybackslash}X@{}}
\toprule
& \textbf{\citet{lu2026duplexsurvey}} & \textbf{This survey} \\
\midrule
Unit coded & Systems & Evidence in reported evaluations \\
Frame & Architectural hierarchy, interaction ontology, state machine & Capability (T, R, G, M) by evidence (conditions, instrument) \\
Tool use & Through one benchmark (FDB-v3) & Core axis G, with state verification and no-action \\
Third parties & Speech to reject or ignore & Addressedness, attribution, authority (M) \\
Finding & Reachable versus realized duplex behavior & Gaps in state checks, real speech, multiparty consequences \\
Output & Taxonomy and system audit & TRG contract and report audits \\
\bottomrule
\end{tabularx}
\caption{Scope relative to the closest contemporary survey.}
\label{tab:lu}
\end{table}

\section{Review Method}
\label{sec:method}
\paragraph{Search and provenance}
The corpus combines five sources under one October 2, 2026 cutoff. Fifteen
supplied seed papers were joined by 20 papers from eight arXiv API queries
(verbatim in \cref{app:search}), screened by title and abstract, and 3
follow-up additions.
Thirty-four named candidates from literature audits, with missing
identifiers resolved through official project pages and exact arXiv acronym
searches, were all retained (33 primary sources and 1 positioning survey).
An evaluation-focused search of logged arXiv queries and citation snowballing
gave \NPool{} further candidates, ordered by a keyword score. From titles and
abstracts in the top 500, an AI assistant proposed \NExpIncluded{} works that
evaluate interactive spoken conversation or its instruments; the list was
accepted after author review, and full-text review excluded \NExpExcluded{}.
The same queries over 24,307 ACL Anthology papers (2024--2026) added 6 works,
selected the same way. Historical searches extended coverage to 2019--2023:
18,058 Anthology papers yielded 5 accepted works; the ISCA Archive search
covered 5,403 Interspeech/SSW papers and yielded 5 accepted works; and a
Semantic Scholar metadata search of ICASSP, SLT, ASRU, and Odyssey yielded 2
accepted works.
\Cref{fig:flow} in \cref{app:search} gives the full selection flow.

\paragraph{Selection and verification}
We retained accessible research papers whose primary contribution bears on
real-time or interactive spoken interaction: architecture, temporal control,
recovery, tool use, robustness, or evaluation validity. Training-data and ASR
studies were retained only where they exposed assumptions of duplex evaluation.
We excluded vendor comparison articles, web pages, unresolvable links, and
unpublished material. Preprint and proceedings versions count as one work.
Each PDF was checked for structure, first-page identity, printed identifier,
and SHA-256 hash, and new arXiv artifacts were version-pinned
(\cref{app:search}). The final set
contains \NArtifacts{} artifacts: \NPrimary{} primary works, \NPositioning{} positioning surveys, and \NMethod{}
methodological source \citep{pedinotti2026structsurvey}. Of the primary works,
\NPrimaryRecent{} are dated 2026 and \NPrimaryArxiv{} were analyzed in their arXiv
versions; \NPrimaryPublished{} have a peer-reviewed record.

\paragraph{Coding protocol}
Each artifact receives a role (evaluation, meta-evaluation, system, data,
analysis, positioning, or method). System, data, analysis, and evaluation
works also receive Axis~A capability codes. Evaluation and
meta-evaluation works, \NEvalMeta{} in total, also receive Axis~B codes for conditions
and instruments plus a one-sentence basis. \Cref{app:codebook} defines every
code. An attribute is coded only when the analyzed PDF states it, so absence
means ``not reported'' rather than ``absent.'' Codes for the \NExpRetained{} works from the evaluation-focused search and the 6
works from the 2024--2026 Anthology search were drafted by the AI assistant with
a verbatim supporting quote each, machine-checked against the cited page, and
accepted after author review; the same procedure applied to the inclusion and
codes of the 12 works from the 2019--2023 ACL, ISCA, and IEEE backfills.
Numerical comparisons stay within their reported conditions.

\paragraph{Auditability}
The companion release (Data Availability) contains the coding file, bases, identifiers, hashes,
search records, and TRG tooling with all audits; coverage
counts are computed from the per-source codes. Source PDFs are not
redistributed.

\section{A Two-Axis Taxonomy}
\label{sec:taxonomy}
An evaluation claim about a voice agent has two parts: \emph{what} capability
was exercised, and \emph{how well} the evidence supports the result. We separate them
(\cref{fig:taxonomy}). The system substrate (speech architecture, training
data, synthesis, and codecs) determines which signals and controls exist; it
is context for evaluation rather than a capability under test. Axis~A lists the capabilities an
interactive agent exercises: timing (T), recovery after disruption (R),
grounded outcomes (G), and, when more than two participants are present,
multiparty conduct (M). Axis~B qualifies any measurement of those capabilities
by its evaluation conditions and its grading instrument.

\begin{figure*}[t]
\centering
\begin{tikzpicture}[font=\small, node distance=2.2mm and 4mm,
  every node/.append style={execute at begin node={\hyphenpenalty=10000\exhyphenpenalty=10000}},
  subbox/.style={draw, rounded corners=2pt, align=center, text width=3.5cm, inner sep=3pt, fill=substrate!12},
  capbox/.style={draw, rounded corners=2pt, align=left, text width=4.3cm, inner sep=3pt, fill=capability!12},
  evbox/.style={draw, rounded corners=2pt, align=left, text width=4.3cm, inner sep=3pt, fill=evidence!14},
  headlbl/.style={font=\small\bfseries, align=center},
  flowarr/.style={-latex, semithick}]
\node[capbox] (T) {\textbf{T: timing}\\prediction, appropriateness, latency boundary};
\node[capbox, below=of T] (R) {\textbf{R: recovery}\\interruption, correction, restored task state};
\node[capbox, below=of R] (G) {\textbf{G: grounded outcome}\\call, execution, verified state, no-action};
\node[capbox, below=of G] (M) {\textbf{M: multiparty} (conditional)\\addressedness, attribution, authority};
\node[draw, dashed, rounded corners=3pt, inner sep=3pt, fit=(T)(M)] (A) {};
\node[headlbl, above=1mm of A] {Axis A: capability exercised};
\node[subbox, left=15mm of A] (S) {\textbf{System substrate}\\end-to-end, cascaded, delegated control; training data; synthesis; codecs};
\node[headlbl, above=1mm of S] {Context};
\node[evbox, right=8mm of A.north east, anchor=north west] (C) {\textbf{Conditions}\\language, real or synthetic speech, acoustics, session length, speakers};
\node[evbox, below=of C] (I) {\textbf{Instruments}\\human reference, model judge, state or trace verification, simulator};
\node[draw, dashed, rounded corners=3pt, inner sep=3pt, fit=(C)(I)] (B) {};
\node[headlbl, above=1mm of B] {Axis B: evidence quality};
\node[draw, thick, rounded corners=2pt, fill=rowtint, align=center, text width=4.3cm, inner sep=3pt, right=8mm of A.south east, anchor=south west] (TRG)
  {\textbf{TRG report}\\T, R, G (and M where applicable), each with conditions and instrument};
\draw[flowarr] (S.east) -- node[above, font=\footnotesize] {constrains} (S.east -| A.west);
\draw[flowarr] (A.east |- B.west) -- (B.west);
\draw[flowarr] (A.east |- TRG.west) -- (TRG.west);
\draw[flowarr] (B.south) -- (B.south |- TRG.north);
\end{tikzpicture}
\caption{Two-axis taxonomy. The substrate constrains which capabilities can be
exercised. Axis~A states what an evaluation measures; Axis~B states the
conditions and grading instrument. TRG asks for every applicable Axis~A result
to be reported with its Axis~B qualifiers.}
\label{fig:taxonomy}
\end{figure*}

\begin{table*}[t]
\centering
\small
\setlength{\tabcolsep}{6pt}
\begin{tabular}{@{}lc|cccc|cccc@{}}
\toprule
& & \multicolumn{4}{c|}{\textbf{Conditions}} & \multicolumn{4}{c}{\textbf{Instruments}} \\
\textbf{Capability} & \textbf{Works} & NE & RS & AV & LS & H & J & S & Sim \\
\midrule
T: timing & 52 & 18 & 24 & 12 & 24 & 28 & 36 & 6 & 10 \\
R: recovery & 35 & 10 & 16 & 10 & 21 & 19 & 29 & 9 & 14 \\
G: grounded outcome & 21 & 6 & 5 & 11 & 14 & 10 & 17 & 13 & 11 \\
M: multiparty & 26 & 10 & 13 & 12 & 12 & 16 & 15 & 2 & 3 \\
 
\bottomrule
\end{tabular}
\caption{Coverage of Axis~A by Axis~B among the \NEval{} evaluation works. Each cell
counts works coding the capability (row) that report the condition or
instrument (column); a work can count in several rows and columns. NE:
non-English; RS: real human speech; AV: acoustic variation; LS: long or
multi-turn session; H: human reference; J: model judge; S: external state or
executed-trace verification; Sim: simulated interlocutor. Counts are generated from the
coding file; codes record only what the analyzed PDF states.}
\label{tab:coverage}
\end{table*}

\Cref{tab:coverage} counts, for the evaluation works, how often each capability
is measured under each condition and instrument. Three gaps stand out. Only
\NState{} of the \NEval{} evaluation works check external state or executed
action traces, all of them grounded-outcome benchmarks, and \NStateSim{} of these
rely on a simulated interlocutor or environment. Only \NGroundedRS{} of the
\NGrounded{} grounded-outcome benchmarks use real human speech: FDB-v3 executes
calls against mock APIs, while WearVox and VGBench match predicted calls
without executing them. Only \NMultiState{} of the \NMulti{} multiparty benchmarks verify state,
and both use only speech not directed at the agent; \NRecoveryAV{} of \NRecovery{} recovery
benchmarks vary acoustics.

\section{System Substrate}
\label{sec:substrate}
\paragraph{Duplex representation is an enabling condition}
Moshi models user and agent speech in parallel codec-token streams
\citep{defossez2024moshi}; Synchronous LLMs ground generation in clock time,
including silence \citep{veluri2024synchronous}. MOSS-Speech studies removing
text guidance without forfeiting pretrained reasoning \citep{mossspeech2025},
and OmniFlatten obtains overlap behavior by post-training an existing backbone
\citep{omniflatten2024}. Hibiki-Zero's simultaneous translation without aligned
supervision \citep{labiausse2026hibikizero} shows why language-dependent
alignment assumptions matter, although translation is not agentic execution.
Such descriptions explain how interaction can be represented without placing
systems in a common performance order. Training data, not only architecture,
constrains realized behavior. DuplexChat constructs speaker-separated
dialogue \citep{duplexchat2026}; ConversationalVoice separates, reconstructs,
and expands real conversations, and explicitly leaves downstream training gains
to future work \citep{he2026conversationalvoice}. Data that preserves overlap
statistics leaves open whether an agent learns context-sensitive floor
management.

\paragraph{Control policy crosses architectural boundaries}
An enterprise tutorial reports that its tested self-hosted end-to-end
configurations did not meet its real-time requirements, while its streaming
cascade achieved 755\,ms time to first audio \citep{qiu2026enterprise}. The
result characterizes those implementations and hosting constraints; it does
not show that end-to-end systems cannot be deployed. DuplexCascade replaces utterance
segmentation with controlled micro-turns \citep{duplexcascade2026};
LTS-VoiceAgent overlaps listening and reasoning through semantic triggering
\citep{zou2026lts}; a duplex frontend can hold the conversation while a text
backend reasons or calls tools \citep{hu2026frontendbackend}. Streaming
transcription \citep{hu2026transcription} makes speech histories more
inspectable, but a transcript remains an imperfect record of what was heard,
rendered, or executed.

\paragraph{Synthesis and codecs bound what can be heard}
Speak While You Think overlaps text and speech generation
\citep{dekel2023speak}; LiveSpeech and CTC-TTS address streaming decoding and
alignment \citep{livespeech2024,ctctts2026}; VoiceChat-TTS targets continuous
interaction with controls for being interrupted \citep{voicechattts2026}.
Codec analysis of Mimi \citep{saloev2026mimi} shows that standard discrimination
metrics miss parts of representation quality, so prosodic claims depend
on what the codec preserves and what an evaluator tests.

\section{Timing: From Prediction to Appropriateness}
\label{sec:timing}
\paragraph{Prediction is not a conversational policy}
VAP predicts joint future voice activity from audio supervision
\citep{ekstedt2022vap}. Prosodic ablations, multilingual transfer, continuous
streaming, and multimodal encoders probe its signal and deployment assumptions
\citep{ekstedt2022prosody,inoue2024multilingual,inoue2024realtime,vapmultimodal2025}.
DualTurn learns conversational dynamics by generative dual-channel pretraining,
then predicts interpretable agent actions \citep{rajaa2026dualturn}, and earable
sensing extends what turn prediction can observe \citep{hamanaka2026sensor}.
Predicting a floor transition is one problem; whether an agent should take it
is a policy question. Controlled ablation of acoustic, prosodic, and semantic
signals finds that adding text can increase premature endpoint detections in
its setting \citep{sharon2026less}. SID-Bench treats an interruption as
something the listener has to understand, not just hear: it labels the
speaker's intent and charges a time cost both for cutting in too early and for
responding too late \citep{xia2026sid}. That does not conflict with the
ablation result above, because spotting where a turn ends and working out what
an overlapping utterance means are two separate jobs. TurnBench reports delay,
recall, and false positives together \citep{turnbench2026}, so a paper that
gives only its fastest setting hides what that speed costs.

\paragraph{Context changes the correct response}
ECHO holds the overlap transcript constant while changing preceding context,
and pair accuracy requires both yielding and floor-keeping decisions to be
correct \citep{zhao2026echo}, so a policy that always stops on overlap cannot
pass. TACT's intent-conditioned scoring uses human floor-transfer timing
distributions \citep{shamsaie2026tact}. Its 9,728 episodes and reported gap
between model and human performance support measuring timing appropriateness:
silence and overlap cannot automatically be classified as errors, although the
measurement depends on intent annotation and the chosen human reference. FLEXI
tests intervention in emergencies \citep{flexi2025}; the floor-taking necessity
study contrasts being invited to speak with recognizing a reason to intervene
\citep{floor2026necessity}. Missing a warranted intervention is also a timing
error. DuplexSpeechBench-IFEval separates deterministic floor adherence from
judge-scored persona content \citep{mathur2026dsbifeval}, and SteerDuplex's
reward analysis illustrates why timing gains must be checked against response
completeness \citep{tyagi2026steerduplex}. Other benchmarks extend timing
evaluation to live human conversation, bilingual, Japanese, and Bangla data,
proactive and instructed behavior, and video calls
\citep{arora2025talkingturns,shen2026codett,fukuda2026m3duplexbench,maruf2026banglaturn,xing2026duplexactbench,tang2026instructfd,mazumdar2026videofdb},
and to late intent revelation and proactive intervention
\citep{xiao2026lateintent,xu2026provoicebench}.

\paragraph{Anticipation and the latency boundary}
Endpoint Anticipation predicts turn completion before the user finishes,
enabling speculative language-model and TTS execution
\citep{endpoint2026anticipation}; it reports a 505\,ms average latency
reduction with a 28.4\% increase in speculative computation. Prediction lead
time, delay after the actual endpoint, and delay to audible playback are not
interchangeable, and aborted speculation must not count as completed work. The
TTS responsiveness benchmark measures latency distributions and intelligibility
across open systems \citep{dinh2025tts}: a low mean can hide long stalls, and
early speech can be incomplete. Reports should identify clock anchors, warm-up,
load, tails, and whether measured audio reached the listener.

\section{Recovery After Disruption}
\label{sec:recovery}
\paragraph{Interruption is an event; recovery is an outcome}
Full-Duplex-Bench scores responses to user interruption \citep{lin2025fdbench};
v1.5 adds overlap types including background speech and talking to others, and
v2 adds multi-turn corrections under an automated examiner
\citep{fdbench15,fdbench2}. MTR-DuplexBench tracks sustained multi-round
conversation \citep{zhang2026mtrduplex}, and the HumDial challenge and its
full-duplex study bring real human-recorded interruptions and rejections into
evaluation \citep{zhao2026humdial,wang2026humdialfd}; FD-Bench, OmniInteract,
and Pardon? add simulated interruptions, nested-context resumption, and repair
\citep{peng2025fdbench,lu2026omniinteract,huang2026pardon}. Such benchmarks establish
whether the agent stops, yields, or continues appropriately; few check
the state of the conversation or task afterward. IHBench makes that question
explicit, evaluating restoration of the workflow, not just prompt stopping, and
finds closed-weight models more robust than open-weight ones \citep{salimi2026ihbench};
EchoChain checks whether continuations after a mid-speech barge-in revise the
task state \citep{modi2026echochain}.

\paragraph{Recovery is an architectural concern}
Voice-Light adds reversible playback and speculative generation, and treats
audio actually played to the user as the authoritative history
\citep{braun2026voicelight}. The Qwen-Audio-Agent
technical report distinguishes speech interruption from task cancellation, and execution completion
from result delivery \citep{deng2026qwenagent}. Stopping audible speech does not
necessarily stop a pending side effect. NemotronLabs VoiceChat reports duplex,
backchannel, and tool-selection results in one model
\citep{nemotron2026voicechat}; \cref{ssec:trg} audits that report. Recovery
evidence should state the disruption, the state that must be restored, and how
that state was checked.

\section{Grounded Outcomes}
\label{sec:grounded}
\paragraph{Three notions of success}
Audio2Tool evaluates tool use from spoken requests of increasing compositional
and acoustic complexity \citep{pahwa2026audio2tool}. VoiceAgentBench tests tool
selection, argument correctness, workflows, and safety in English and Indic
languages with more than 6,000 synthetic spoken queries in its analyzed version
\citep{jain2025voiceagentbench}, and FDB-v3 adds natural disfluency
\citep{lin2026fdbv3}. Between them, these benchmarks tell a plausible answer
apart from a correct structured call, but a correct call still says nothing
about what happened downstream. $\tau$-Voice puts grounded domain tasks into
full-duplex interaction \citep{ray2026tauvoice}, and VAmoS seeds a backend with
data and then checks assertions against both the interaction trace and the
database state \citep{meyer2026vamos}.
EVA-Bench uses validated bot-to-bot simulation and repeated trials to
separate peak from reliable performance \citep{bogavelli2026evabench}.
APEX-Voice, DuplexWorld, and $\tau$-Elicitation, a 200-task benchmark of how
well voice agents capture key entities, check the final environment, world, or
database state, all with simulated users
\citep{mathur2026apexvoice,bhosale2026duplexworld,ray2026tauelicitation}.
Talk2Agent and spoken versions of tool-calling benchmarks ask a different
question: how much of what a text agent needs survives the move to voice
\citep{chiba2026talk2agent,laskar2026texttovoice}. Perceived fluency, correct
calls, policy compliance, and verified state are four different kinds of
evidence, and a report should give each one separately before combining them.

\paragraph{The correct outcome may be no action}
VGBench uses one action space (silence, tool call, or spoken response) across
side-talk, self-talk, and speaker-switch conditions \citep{zhang2026vgbench}.
Its paired speaker-switch controls hold specified words fixed while changing
several acoustic-context cues, which the authors describe as multi-cue gating
rather than isolated speaker identity. A correct tool choice can accompany a
wrong decision to act. Grounded evaluation therefore needs an action-eligibility
rule and a trace showing that the world appropriately remained unchanged. A mute
rate alone does not show that no background task executed.

\paragraph{Locating the failure}
MTVA-Bench isolates the language model inside a noisy cascade
\citep{mishra2026mtva}; SpeechGym traces how misheard argument values defeat
otherwise correct tool reasoning, and reports an unauthorized write under an
insistent caller \citep{speechgym2026}. Outcome scoring identifies failure;
component and process evidence is needed to locate it.

\section{Multiparty Conduct}
\label{sec:multiparty}
MP-Bench measures participation among several speakers \citep{shih2026mpbench}.
Duplex-MPE separately assesses response initiation, answer accuracy, silence
preservation, and stopping in multiparty audio \citep{duplex2026mpe}. Neither
frequent responding nor frequent silence is inherently good: false and missed
interventions need separate denominators. Triadic VAP and MuVAP extend turn
prediction beyond two participants \citep{elmers2025triadic,qi2026muvap}, and
Adaptive Turn-Taking conditions a multiparty agent on its assigned role
\citep{mitra2026adaptive}. Speaker-attributed HEAR uses counterfactual voice
grounding to diagnose reasoning about who spoke \citep{lee2026hear}; MSI-Bench
adds speaker-scoped memory, instructions, and expected tool calls
\citep{xiong2026msibench}. TPI-Bench interrupts a user with a third speaker
\citep{lee2026tpibench}. WearVox tests side-talk rejection and
tool calls on smart-glasses recordings \citep{lin2026wearvox}. Speaker-centric,
addressee, and social-interactivity benchmarks widen the measurement of who
speaks to whom
\citep{sun2026msubench,kwon2025m3slu,inoue2025addressee,fukuda2026meetingturntaking,xie2026socialomni}.
Identifying a speaker does not establish that the speaker may issue a command
or receive a private answer (\cref{tab:coverage}).

\section{Evidence Quality}
\label{sec:evidence}
\paragraph{Language and acoustics are conditions}
\citet{tau2026multilingual} extend grounded voice tasks to five languages
beyond English in $\tau$-Multilingual and report language-dependent failure
patterns. IndicFDB tests conversational behavior in ten languages spoken in
India, using VAD-based timing where word-aligned ASR is unreliable, and warns
that mined events differ across languages \citep{indic2026fdb}. French FDB
contrasts transferable timing metrics with language-sensitive content
assessment \citep{french2026fdb}, which bears on metric validity as well as
model transfer. VoiceBench and VCB Bench vary voice, environment, and content
with different combinations of real and synthetic speech
\citep{chen2024voicebench,hu2025vcb}. EVA-Bench and $\tau$-Voice apply accent
and noise conditions \citep{bogavelli2026evabench,ray2026tauvoice}; IRAF studies
interfering speech in the user channel \citep{zhong2026iraf}. MMedFD's
healthcare sessions carry overlap, timing, and role annotations
\citep{chen2025mmedfd}, but recognizing clinical content does not show that an
action based on it would be safe. The demographic-bias HEAR benchmark contains
87k recordings from 843 participants \citep{yan2026hearbias} and should not be
confused with speaker-attributed HEAR. SpeechJBB finds that spoken jailbreak
vulnerability under code switching cannot be reduced to comprehension failure
\citep{ceccatelli2026speechjbb}. Further component benchmarks cover multi-turn,
in-the-wild, long-history, and role-play spoken dialogue (\cref{tab:map-component}).

\paragraph{The grading instrument is part of the result}
Human raters provide contextual reference judgments, but their instructions and
inter-rater agreement must be reported. In the voice-agent judge study
\citep{purwar2026judges}, model-judge reliability depends on the metric and
configuration, so a correlation for one rubric does not carry over to every
safety or recovery metric. Instruments are now studied directly, from audio and
safety judges and their acoustic shortcuts
\citep{sayyad2026lalmjudges,ivry2026lalmasajudge,huo2026speechjudgeaudit} to
comparisons of static benchmark scores with user preferences \citep{li2025mindthegap} and of
reference-based or behavioral scores with human conversation
\citep{bokkahallisatish2026cds,hallur2026prosodyref,zhang2026turnnat,inoue2024userbehaviors}.
Deterministic checks answer narrow questions reproducibly; state assertions
check consequences, and traces connect them to actor, action, and sequence. A
judge should not replace a missing state verifier without saying so. Paired
designs (ECHO's decisions, HEAR's speaker counterfactuals, VGBench's acoustic
controls) limit the chance of succeeding through constant policies or semantic
priors \citep{zhao2026echo,lee2026hear,zhang2026vgbench}. Simulation adds
assumptions: EVA-Bench validates and regenerates erroneous simulated
conversations \citep{bogavelli2026evabench}, and $\tau$-Voice decouples
simulation from wall-clock execution \citep{ray2026tauvoice}. Reports should
attribute each failure to the agent or the simulator and keep composite
components visible.

\begin{table*}[t]
\centering
\footnotesize
\setlength{\tabcolsep}{4pt}
\begin{tabularx}{\textwidth}{@{}l>{\raggedright\arraybackslash}X>{\raggedright\arraybackslash}X>{\raggedright\arraybackslash}X@{}}
\toprule
\textbf{TRG item} & \textbf{NemotronLabs VoiceChat} (end-to-end, tools) & \textbf{Frontend-backend} (hybrid delegation) & \textbf{Moshi} (end-to-end, no tools) \\
\midrule
T latency & 0.448\,s / 0.480\,s (FDB 1.0); statistic and anchors unstated & Median 3.14\,s, end of user speech to non-filler response (EVA-Bench); tails not reported & 160\,ms theoretical, 200\,ms in practice; statistic, anchors, and hardware unstated \\
T appropr. & Pause takeover 15.3\% / 25.5\% & Candor pause takeover 68.2\%, against 53.2\% without tool training & Pause, gap, and overlap totals in self-generated dialogue against ground truth \\
R & Takeover 100\%, resume after backchannel 93\%; task state not measured & Interruption takeover 89.5\%, barge-in accuracy 99\%; task state not reported & Not evaluated \\
G & Tool-selection F1 82.5\%, Pass@1 33.0\%; call matching, no external state & EVA-Bench task completion 57.3\%, defined on database state (report says GPT-5.2 judged all metrics); caller receives agent text & No action space \\
G no-action & Not reported & Up to 81.2\% of irrelevant calls rejected; no action trace & Not applicable \\
M & Dyadic only; justified & Dyadic only; justified & Dyadic only; justified \\
\bottomrule
\end{tabularx}
\caption{TRG version-2 audits of three published reports
\citep{nemotron2026voicechat,hu2026frontendbackend,defossez2024moshi}. Values
are transcribed from the reports (for the frontend-backend report, its
largest-backend configuration). The validator marks all three non-compliant;
it checks disclosure, not system quality.}
\label{tab:audit}
\end{table*}

\section{Three Claims and a Reporting Contract}
\label{sec:analysis}
\paragraph{Claim 1: architecture is a deployment constraint, not a verdict}
Cascaded duplex control, speculative scheduling, and frontend-backend delegation
show that duplex interaction is not confined to one architectural family
\citep{duplexcascade2026,endpoint2026anticipation,hu2026frontendbackend}. A
cascade can still lose paralinguistic information, and latency results remain
tied to hardware and hosting. EVA-Bench's multi-architecture assessment finds
no evaluated system above 0.5 on both accuracy (EVA-A) and experience (EVA-X) pass@1
\citep{bogavelli2026evabench}. Systems should be compared under their actual
hosting, control, signal, and execution constraints. A reachable duplex state
is a capacity; competence has to be observed.

\paragraph{Claim 2: benchmarks can verify consequences, but few do}
Backend assertions and grounded tasks provide stronger outcome evidence than
an agent's own report of completion \citep{ray2026tauvoice,meyer2026vamos}, and recovery
and eligibility diagnostics expose further ways an apparently successful call
can be wrong \citep{salimi2026ihbench,zhang2026vgbench}. Yet only \NState{} of \NEval{}
evaluation works check external state or executed action traces (\NStatePublished{} of
the \NEvalPublished{} with a peer-reviewed record).
Of the \NState{}, \NStateRecent{} date from 2026, as do \NPrimaryRecent{} of \NPrimary{}
primary works, so the corpus shows that such evaluation now exists; it cannot
establish a field-wide temporal trend. State verifiers also need costly
domain environments and policies. Reports should therefore keep prediction,
execution, and consequence distinct and include cases in which no action is
warranted.

\paragraph{Claim 3: the dyadic assumption is breaking down}
MP-Bench, MSI-Bench, Duplex-MPE, TPI-Bench, WearVox, and speaker-attributed HEAR measure
participation, attribution, third-party speech, and speaker-scoped reasoning beyond one user and
one agent \citep{shih2026mpbench,xiong2026msibench,duplex2026mpe,lee2026tpibench,lin2026wearvox,lee2026hear},
and necessity-sensitive evaluation shows that silence can also fail
\citep{floor2026necessity}. The evidence is still sparse: \NMulti{} of \NEval{}
evaluation works code multiparty conduct, and only \NMultiState{} verify the
resulting state. Role-conditioned models address part of the problem
\citep{mitra2026adaptive}, but knowing who is speaking does not settle who
may authorize an action. Deployments should specify who may address the
agent, issue actions, and receive information.

\subsection{TRG: Timing-Recovery-Grounded}
\label{ssec:trg}
TRG specifies what a report must disclose, not how to design a benchmark,
choose metrics, or rank systems. Each letter is an Axis~A capability, and every
reported result must state its Axis~B conditions and grading instrument.
\begin{enumerate}[leftmargin=1.4em,itemsep=1pt,topsep=2pt]
\item[\textbf{T}] \emph{Timing.} Report latency with its start and end points,
unit, statistic, and test conditions, plus tail latency or a note that it was
not measured. If the agent decides when to take the floor, also
report whether that timing suited the context or the user's intent, and how
this was annotated and scored \citep{xia2026sid,shamsaie2026tact}.
\item[\textbf{R}] \emph{Recovery.} Name the disruption and measure, separately
from undisturbed performance, whether conversational and task state were
restored. Stopping audio, resuming speech, and restoring the workflow are
separate outcomes \citep{salimi2026ihbench,braun2026voicelight}.
\item[\textbf{G}] \emph{Grounded outcome.} Verify from external evidence, not
the agent's own report, that the authorized target state was reached. When no
action is warranted, state the eligibility rule and show from action traces
that external state did not change \citep{meyer2026vamos,zhang2026vgbench}.
\item[\textbf{M}] \emph{Multiparty (conditional).} With more than two legitimate
participants, report whether the agent spoke and stayed silent when warranted,
attributed speech where relevant, and disclosed information or acted only as
authorized. For a dyadic deployment, state why M does not apply.
\end{enumerate}
Language, accent, noise,
speaker configuration, and system load are conditions reported under every
letter, not extra letters. Compliance means the report is complete; it says
nothing about performance or safety.

\paragraph{Worked audits}
We applied the version-2 template and validator, without re-evaluating any
system, to three reports that differ in architecture and tool use
(\cref{tab:audit}). In the NemotronLabs report, tool
routing (82.5\% F1) is much stronger than complete calls (33.0\% Pass@1, which
requires exact tools and all arguments), no tool result checks external state,
and barge-in is unavailable while a tool executes. The frontend-backend report
comes closest: it anchors a median latency and reaches state-defined task
completion through EVA-Bench, but its EVA runs return agent text to the
simulated caller, and tool-call training raised pause takeover from 53.2\% to
68.2\%. Moshi's report supports conversational but not agent claims,
since it evaluates neither recovery nor actions. None of the three measures
task state after interruption. Audits of 10 randomly sampled
reports are all non-compliant; none verifies external state
(\cref{app:trgsample}).

\section{Open Challenges}
\label{sec:challenges}
\paragraph{Grounded tasks from real speech}
Among grounded benchmarks with real speech, only FDB-v3 executes calls, against
mock APIs \citep{lin2026fdbv3}. Replaying recorded spoken requests
against a state verifier would show how often SpeechGym's misheard-argument
failures \citep{speechgym2026} occur with human speakers.

\paragraph{Recovery during pending actions}
Neither tool-using report in \cref{tab:audit} interrupts a pending call; an
evaluation should, and verify whether the action completed, was cancelled, or
was misreported.

\paragraph{Timing under cost}
Operating curves of latency, inappropriate intervention, and
speculative cost would make endpoint anticipation and
SID-Bench comparable \citep{endpoint2026anticipation,xia2026sid}.

\paragraph{Multiparty authority with consequences}
MSI-Bench specifies expected tool calls
per speaker \citep{xiong2026msibench}. Adding an authorization and
disclosure policy, then verifying the action and what each participant was
told, would carry grounded evaluation beyond the dyad.

\section{Conclusion}
\label{sec:conclusion}
We reviewed \NPrimary{} primary sources and argue that real-time voice agents
should be evaluated as policies acting in time, in conversation, and on an
external environment. The two-axis coding locates three gaps: few evaluations verify external
state, grounded tasks rarely use real human speech, and multiparty evaluation
rarely checks consequences.
From these findings we derive TRG, a minimum reporting contract for
context-appropriate timing, restored conversation and task state, and
externally verified outcomes, each with its conditions and grading method.

\section*{Limitations}
Most of the corpus is recent: \NPrimaryRecent{} of the \NPrimary{} primary
works are dated 2026, and most have no peer-reviewed version yet. The search is
not exhaustive. It covers nothing before 2019, the Interspeech 2026 archive
could be checked only through public indexes, and 37 IEEE candidates had no
open full text. Candidates from the additional searches and from a second round
of citation tracking were not screened for inclusion and lie outside the corpus
(\cref{app:rankextension}). The corpus
therefore cannot show how evaluation practice has changed over time, and some
relevant work is probably missing.

No second person checked the screening. An AI assistant proposed which works to
include and drafted their codes, and one author accepted each batch. A code
records only what a paper states explicitly, so a work that did something
without reporting it is coded as not doing it. We did not rerun any experiment,
and we do not pool results, because the works differ in datasets, evaluation
protocols, and reporting practices. Sensitivity analyses by publication status,
publication year, and search source all show that external state verification
remains uncommon (\NSensPeerState{} of \NSensPeerEval{}, \NSensPreState{} of
\NSensPreEval{}, and \NSensBaseState{} of \NSensBaseEval{} evaluation works;
\cref{app:sensitivity}), although they cannot rule out selection bias.

The TRG audits illustrate the contract; they do not test it. We chose the three
worked audits to contrast architectures, and each of the ten sampled audits
covers a single report. They show which gaps TRG exposes, not whether adopting
it improves reproducibility or deployment decisions, and we have not measured
how much work it adds for authors. TRG is also not a safety, security,
fairness, or privacy framework. No independent external adoption has been
verified here. A common-condition multi-system study with matched clean and
interrupted episodes, fixed client-playout clock anchors, externally checked
state and action traces, and case-level paired uncertainty would test whether
latency-only comparisons conceal recovery or task-completion differences.
Such a study remains future work, not a result of these audits.

\section*{Ethical Considerations}
This review analyzes published work and involves no new human-subject data
collection. Source PDFs are hashed and not redistributed because most carry no
redistribution grant. Evaluation gaps have concrete stakes: misattributed
speakers can expose private information, unverified tool calls can produce
unauthorized side effects, and English or clean-audio evaluation can hide
failures for other languages, accents, and speakers. TRG makes such omissions
visible, but a compliant report can still describe an unsafe system. The corresponding author (S.~Negi) used Anthropic's Claude for literature retrieval and corpus assembly; for
ranking and proposing candidates from the evaluation-focused and Anthology
searches; for programmatic verification of source PDFs; for proposing
taxonomy codes with supporting quotes; for drafting the sampled TRG audits;
and for drafting and revising text. The authors checked this material against
the primary sources and take full responsibility for the content.

\section*{Data Availability}
The matching 140-artifact companion corpus and TRG tooling are available in
release v1.2.0 at
\url{https://github.com/shivamnegi92/voice-agent-eval-corpus/releases/tag/v1.2.0}.
The Zenodo concept record
(\href{https://doi.org/10.5281/zenodo.22924302}{doi:10.5281/zenodo.22924302})
identifies the archive family; use the version-specific release and checksums
rather than assuming the concept record already contains this revision.
Metadata records identifiers, analyzed PDF hashes, verification dates, source
roles, and the two-axis taxonomy codes from which \cref{tab:coverage} is
generated. Source PDFs remain private and are not redistributed.

\appendix

\section{Search Record and Selection Flow}
\label{app:search}
The eight arXiv API queries were the following; candidates were ranked by
relevance and screened by title and abstract.
\begin{itemize}[leftmargin=1.2em,itemsep=0pt,topsep=2pt,font=\small]
\small
\item \texttt{ti:"Voice Activity Projection"}
\item \texttt{abs:"voice agent" AND abs:"benchmark"}
\item \texttt{abs:"full-duplex" AND abs:"spoken dialogue"}
\item \texttt{abs:"speech-to-speech" AND abs:"end-to-end"}
\item \texttt{abs:"neural audio codec" AND abs:"speech"}
\item \texttt{abs:"barge-in" OR abs:"interruption handling"}
\item \texttt{abs:"streaming" AND abs:"text-to-speech" AND abs:"latency"}
\item \texttt{all:"Full-Duplex-Bench"}
\end{itemize}
Named candidates were resolved through supplied candidate audits, official
project pages, and exact arXiv searches for \texttt{IRAF} and \texttt{HEAR
demographic bias}, all recorded on October 2, 2026. The evaluation-focused
search, run on October 5, 2026, used 22 logged arXiv queries, three of them
added after a recall check had retrieved 63 of 66 testable corpus works, plus
backward snowballing (arXiv identifiers in the reference lists of the corpus
evaluation works) and forward snowballing (Semantic Scholar citations of those
works). A transparent keyword score ordered the pool, placing 31 of the 36
evaluation works found by the other sources in its top 400. General audio
understanding, ASR-only, TTS-only, translation, and text-only agent benchmarks
were out of scope; full-text review excluded a streaming video benchmark with
text queries and a text-only benchmark. The queries with run date and hit
counts, the recall check, the snowball records, the ranking procedure, and the
screening and coding records are in the supplement. Identity checks caught one
real error: a circulated identifier for FDB-v3, 2602.05105, resolves to an
unrelated robotics paper; the verified identifier is 2604.04847.
The Anthology search, run on October 6, 2026, applied regular-expression
equivalents of the same queries to titles and abstracts from the 2024--2026
ACL, EMNLP, NAACL, EACL, Findings, TACL, CL, COLING, AACL/IJCNLP, LREC, SIGDIAL,
and IWSDS volumes. Of 170 matches, 13 were already in the corpus and 37 had been
screened within the arXiv pool's top 500; title and abstract screening of the
remaining 120 included 6, whose PDFs were retrieved from the Anthology and
identity-checked. A historical Anthology search over the same venues for
2019--2023 covered 18,058 papers, matched 74 (one already in the corpus), and
screened 73 new candidates by title and abstract; five proceeded to full text
and were accepted. IEEE speech venues (ICASSP, SLT, ASRU, Odyssey) were searched
through Semantic Scholar metadata: 231 records, 115 query matches, and 79 new
candidates. Two with arXiv versions were screened at full text first and both
were accepted; a later full-text pass read 40 of the remaining 77 open copies
and accepted none, leaving 37 unscreened beyond metadata because IEEE Xplore
has no free bulk full-text API. The ISCA Archive
historical search covered 5,403 papers from Interspeech 2019--2023 and SSW
2019/2021/2023; 108 matched (one already in the corpus, one in the arXiv pool),
106 were new. Three were first screened at full text and accepted; a later
full-text pass read 102 of the remaining 103 (one paper was unavailable) and
accepted two more evaluation works. The 2019--2023
searches therefore improve historical coverage but do not amount to exhaustive
screening of those venues. The Interspeech 2026 archive was not available for automated retrieval; an index probe (Crossref,
OpenAlex, Semantic Scholar, arXiv) found 574 Interspeech 2026 records, 42
regex matches, and 3 not seen before, provisionally triaged as 2 ASR-only
exclusions and 1 uncertain. These three were not screened for inclusion and lie
outside the corpus. A retry found no open
copy for any of the 37 IEEE candidates. Anthology and ISCA records also give
published versions for corpus works analyzed as preprints; their bibliography
entries cite the published version.

\begin{figure}[t]
\centering
\begin{tikzpicture}[font=\small, node distance=2.6mm,
  every node/.append style={execute at begin node={\hyphenpenalty=10000\exhyphenpenalty=10000}},
  flowbox/.style={draw, rounded corners=1.5pt, align=center, text width=4.1cm, inner sep=2.5pt},
  flowout/.style={draw, align=center, text width=2.4cm, inner sep=2.5pt, fill=rowtint},
  flowarr/.style={-latex, semithick}]
\node[flowbox] (seed) {Seed collection: 51 entries};
\node[flowout, right=3mm of seed] (excl) {Excluded: 17 web, 18 vendor, 1 unavailable};
\node[flowbox, below=of seed] (seedp) {15 seed research papers};
\node[flowbox, below=of seedp] (query) {+ 20 from arXiv queries\\+ 3 follow-up additions\\+ 3 positioning or method};
\node[flowbox, below=of query] (sub) {41 artifacts\\(38 primary; initial pool)};
\node[flowbox, below=of sub] (screen) {+ 34 named candidates screened, all retained (33 primary, 1 positioning)};
\node[flowbox, below=of screen] (evalsearch) {+ Evaluation-focused search: \NPool{} ranked candidates; \NExpIncluded{} proposed and accepted; \NExpExcluded{} excluded at full text};
\node[flowbox, below=of evalsearch] (acl) {+ 2024--2026 ACL Anthology: 6 works\\+ 2019--2023 ACL, ISCA, IEEE: 12 works};
\node[flowbox, below=of acl, fill=rowtint] (final) {\textbf{\NArtifacts{} artifacts}: \NPrimary{} primary, \NPositioning{} positioning, \NMethod{} method};
\node[flowbox, below=of final] (code) {Role for every artifact; two-axis codes where applicable};
\draw[flowarr] (seed) -- (excl);
\draw[flowarr] (seed) -- (seedp);
\draw[flowarr] (seedp) -- (query);
\draw[flowarr] (query) -- (sub);
\draw[flowarr] (sub) -- (screen);
\draw[flowarr] (screen) -- (evalsearch);
\draw[flowarr] (evalsearch) -- (acl);
\draw[flowarr] (acl) -- (final);
\draw[flowarr] (final) -- (code);
\end{tikzpicture}
\caption{Source selection. \NPrimaryPublished{} primary works cite
peer-reviewed versions instead of preprints, without changing the count.}
\label{fig:flow}
\end{figure}

\section{Codebook}
\label{app:codebook}
\paragraph{Roles}
\emph{Evaluation}: proposes a benchmark or protocol scoring system behavior.
\emph{Meta-evaluation}: assesses an evaluation instrument. \emph{System},
\emph{data}, and \emph{analysis} works receive Axis~A codes only;
\emph{positioning} and \emph{method} sources receive no codes.
\paragraph{Axis A}
C (component): speech or language quality without interaction timing, such as
spoken question answering or safety. T: when to speak, yield, or stay silent,
or latency. R: behavior after interruption, correction, or other disruption.
G: tool calls, task completion, or external state. M: more than two speakers,
addressedness, or speaker attribution.
\paragraph{Axis B}
NE: any evaluation language other than English. RS: evaluation audio recorded
from humans rather than synthesized. AV: controlled acoustic, accent, room, or
speaker perturbation. LS: multi-turn or long-session interaction. H: human
ratings or annotations used as a reference or to validate another instrument.
J: a language-model judge. S: assertions on external state or executed action
traces; execution against a mock or simulated backend counts, while matching
predicted tool calls against references without executing them is not S. Sim: a simulated interlocutor or environment.

\section{Sampled TRG Audits}
\label{app:trgsample}
We drew 10 evaluation works at random (fixed seed, recorded in the supplement)
and filled in a TRG version-2 report from each paper's own text for the first
system in its main results table, without re-evaluating anything; the AI assistant drafted the reports and
the validator graded them. All 10 are non-compliant. None verifies an outcome
against external state; 9 fail a timing requirement (latency anchors, units, or
appropriateness); 8 omit at least one evaluation condition; 6 fail recovery; and
all 3 with more than two speakers lack an intervention measure. Many of these
benchmarks never claim agent outcomes, so the audits show what their reports
cannot support, not that the benchmarks are badly designed. Fields containing placeholders such as ``not reported'' are treated as missing.

\section{Rank-Extension and Vocabulary Probes}
\label{app:rankextension}
A title/abstract screen of evaluation-focused search ranks 501--1000 surfaced
53 candidates. Provisional full-text review of all 53 classified 36 as possible
evaluation works, 16 as possible other works, and 1 as out of scope; these are
not inclusion decisions. Full text was available for every candidate in this
review. The supplementary materials provide the records, supporting quotations, and review labels.

The supplementary 2019--2023 vocabulary probe matched 152 ACL and 185 ISCA
records; 92 ACL and 118 ISCA records were absent from prior search files. A
title/abstract triage of these absent records identified 15 possible-evaluation
and 8 uncertain ACL records, and 4 possible-evaluation and 12 uncertain ISCA
records. They fall within the 2019--2023 backfill period but were not screened
for inclusion and lie outside the corpus. The fresh IEEE vocabulary query could not be
run because network access was blocked; cached
IEEE metadata was already filtered using the original terms. These probes test
query sensitivity, not historical recall or completeness. Full queries,
records, and screening decisions are provided in the supplementary materials.

\section{Sensitivity of the Coverage Counts}
\label{app:sensitivity}
\Cref{tab:sensitivity} recomputes the three headline ratios for subsets of the
evaluation works. Later additions are the works included by the
evaluation-focused, ACL Anthology, IEEE and ISCA Archive searches. State
verification stays rare in every subset, including pre-2026 work: only
\NSensPreState{} of \NSensPreEval{} evaluation works published before 2026
verify state. The real-speech ratio does not hold up the same way: it rises to
\NSensPeerGroundedRS{}/\NSensPeerGrounded{} for peer-reviewed and
\NSensPreGroundedRS{}/\NSensPreGrounded{} for pre-2026 works, but on so few
grounded benchmarks that it cannot be tested there. The pre-2026 subset is also
under-sampled rather than representative, because the historical searches
reused recent terminology and started in 2019.
Denominators below 10 are too small to read as rates.

\begin{table}[h]
\centering
\small
\setlength{\tabcolsep}{3pt}
\begin{tabular}{@{}lrlll@{}}
\toprule
Subset & Eval & S & G with RS & M with S \\
\midrule
All works & 91 & 13/91 & 5/21 & 2/26 \\
Peer-reviewed & 31 & 3/31 & 2/3 & 1/8 \\
Before 2026 & 31 & 2/31 & 3/4 & 0/7 \\
No targeted additions & 36 & 7/36 & 2/12 & 2/12 \\
 
\bottomrule
\end{tabular}
\caption{Headline ratios by subset of evaluation works, generated from the
coding file. S: state verification; G: grounded outcome; RS: real speech; M:
multiparty.}
\label{tab:sensitivity}
\end{table}

\section{Per-Work Coding}
\label{app:coding}
Tables~\ref{tab:map-timing}--\ref{tab:map-component} list every evaluation and
meta-evaluation work, grouped by the capability it primarily targets, with its
citation, Axis~A and Axis~B codes, and what it scores. They are generated from
the coding files, and the supplement records a one-sentence coding basis and a
verbatim supporting quote for each row.

\begin{table*}[p]
\centering
\footnotesize
\setlength{\tabcolsep}{3.5pt}
\begin{tabularx}{\textwidth}{@{}p{4.6cm}lllX@{}}
\toprule
\textbf{Work} & \textbf{A} & \textbf{Cond.} & \textbf{Instr.} & \textbf{What is scored} \\
\midrule
BanglaTurn \citep{maruf2026banglaturn} & T & NE, RS & H & Binary endpoint classification accuracy and false interrupt/wait rates on Bangla clips \\
CoDeTT \citep{shen2026codett} & T, M & NE, RS, AV, LS & -- & Accuracy of turn-taking actions and intent errors across 14 scenarios \\
DSB-IFEval \citep{mathur2026dsbifeval} & C, T, R & LS & H, J, Sim & Deterministic floor adherence; judged persona content \\
DuplexAct-Bench \citep{xing2026duplexactbench} & C, T, R & NE, RS & J & Whether, when and how systems interrupt, yield, stay silent, initiate or backchannel \\
ECHO \citep{zhao2026echo} & T, M & NE, LS & H & Paired yield and keep decisions under changed preceding context \\
FLEXI \citep{flexi2025} & T, R & -- & J & Full-duplex interaction in synthesized scenarios, including model interruption in emergencies \\
Floor necessity \citep{floor2026necessity} & C, T & -- & J & Speaking when invited versus when a reason to intervene arises \\
French FDB \citep{french2026fdb} & T, R & NE, RS & H, J & Timing metrics and judged content in Canadian and European French \\
FTAD fine-grained turn-taking prediction benchmark \citep{chen2021ftad} & T, R & NE, RS & -- & Classification accuracy of a context-aware self-attention baseline (and LSTM-CRF for backchannel) on EOT, respond-time, break and backchannel prediction, FTAD-sw vs FTAD-zh, with and without dialogue context \\
Game-Time \citep{chang2025gametime} & C, T & -- & H, J & Instruction following under duration, silence, tempo and synchronized-overlap constraints \\
Incremental ASR/diarization evaluation (IBM, Google, Microsoft) \citep{addlesee2020incremental} & T, M & RS & -- & First-Occurrence and Final-Decision latency, Word Survival Rate (stability), disfluency-preserving WER gain, overlap WER gain, diarization error rate \\
Instruct-FD \citep{tang2026instructfd} & T, R & LS & H, J, Sim & Judge pass rate for following instructed turn-taking behavior per instruction type \\
LateIntent-Bench \citep{xiao2026lateintent} & C, T & -- & J & Premature response onset before intent reveal, plus harmful and benign engagement rates \\
M3-DuplexBench \citep{fukuda2026m3duplexbench} & C, T, R & NE, RS, LS & J & Turn-taking, pause, backchannel and barge-in timing plus judged relevance, consistency, QA accuracy \\
OmniInteract \citep{lu2026omniinteract} & T, R & NE, RS & H, J & Answer quality within valid response windows, interruption handling, and context resumption \\
On-device user-perceived latency (UPL) benchmark for streaming E2E ASR \citep{shangguan2021latency} & C, T & RS & H & P50/P90 user-perceived latency, endpointer lag, decoder catchup lag and first-token delay measured on-device, plotted against WER, across model architecture, stride, training loss and endpointer (StaticEP, NEP, E2E-EP) \\
ProVoice-Bench \citep{xu2026provoicebench} & T, G, M & AV & J & Intervene-or-stay-silent decisions (accuracy, FPR, recall) plus judged tool-call/response accuracy \\
SID-Bench \citep{xia2026sid} & T & NE, RS, AV & -- & Semantic interruption intent; temporal cost of false alarms and late responses \\
TACT \citep{shamsaie2026tact} & T & RS, LS & H & Intent-conditioned timing against human floor-transfer distributions \\
Talking Turns \citep{arora2025talkingturns} & T, R & RS, LS & H & Timing of speaking up, backchannels, interruptions, floor cues and yielding \\
Timing Generating Network (TGN) turn-taking evaluation \citep{fujie2021timing} & T, M & RS & -- & Turn-taking timing prediction accuracy/appropriateness versus a conventional hard-decision (VAD + response-time estimation) baseline \\
TTS responsiveness \citep{dinh2025tts} & C, T & -- & -- & Latency distribution, tail latency, and intelligibility of open TTS systems \\
Turn-taking prediction on EgoCom \citep{fatan2024threeM} & T, M & RS & -- & Next-speaker prediction accuracy (no one / host / one of two participants), compared to existing baselines and alternative transformer architectures \\
TurnBench \citep{turnbench2026} & T & RS, LS & H & Delay, recall, and false positives on recorded dyadic conversation \\
VAP-based turn-taking cue evaluation of TTS systems \citep{ekstedt2023automatic} & T & -- & J & Predicted next-speaker probability (VAP) at turn-hold and turn-yield points, compared across five TTS systems and prosody conditions \\
VideoFDB \citep{mazumdar2026videofdb} & T & RS & H, J & Turn timing and nonverbal-cue appropriateness of agent responses \\
Within-turn TRP prediction (ICC) \citep{umair2024whentospeak} & T & -- & H & Precision, recall, F1, free-marginal kappa and temporal distance of predicted TRPs versus participant-agreed TRPs \\
\bottomrule
\end{tabularx}
\caption{Timing: the 27 works whose primary target is timing. Codes as in \cref{app:codebook}; ``--'' means none reported.}
\label{tab:map-timing}
\end{table*}

\begin{table*}[p]
\centering
\footnotesize
\setlength{\tabcolsep}{3.5pt}
\begin{tabularx}{\textwidth}{@{}p{4.6cm}lllX@{}}
\toprule
\textbf{Work} & \textbf{A} & \textbf{Cond.} & \textbf{Instr.} & \textbf{What is scored} \\
\midrule
Duplex Conversation interruption-handling A/B test \citep{jin2021duplex} & R & RS & -- & Intention recognition rate and dialogue completion rate, with the new interruption/discontinuous-expression detection on vs off, on live outbound calls \\
EchoChain \citep{modi2026echochain} & R & LS & H, J, Sim & Correct task-state revision after mid-speech interruptions, per conversation rubric \\
FD-Bench \citep{peng2025fdbench} & T, R & AV, LS & J, Sim & Interruption success rates, response delays, noise interrupts and judged reply quality \\
FDB-v1.5 \citep{fdbench15} & T, R, M & AV & J & Responses to interruption, backchannel, talking to others, and background speech \\
FDB-v2 \citep{fdbench2} & C, T, R & LS & H, J, Sim & Multi-turn behavior, including corrections, under an automated examiner \\
Full-Duplex-Bench \citep{lin2025fdbench} & T, R & RS & H, J & Pause handling, backchanneling, turn-taking, and user interruption \\
HumDial challenge \citep{zhao2026humdial} & C, T, R, M & NE, RS, LS & H, J & Interruption and rejection handling in authentic conversations \\
HumDial study \citep{wang2026humdialfd} & T, R, M & NE, RS & J & Interruption, rejection, and third-party speech in real recordings \\
IHBench \citep{salimi2026ihbench} & R, G & LS & H, J, Sim & Restoration of the task workflow after interruption \\
IndicFDB \citep{indic2026fdb} & T, R & NE, RS & H, J & Conversational behavior in ten languages spoken in India, with VAD-based timing \\
MTR-DuplexBench \citep{zhang2026mtrduplex} & C, T, R & RS, LS & J & Behavior sustained over multi-round full-duplex conversation \\
Pardon? \citep{huang2026pardon} & C, R & AV & J & Answer accuracy on answerable audio and conversational repair on unanswerable audio \\
\bottomrule
\end{tabularx}
\caption{Recovery after disruption: the 12 works whose primary target is recovery after disruption. Codes as in \cref{app:codebook}; ``--'' means none reported.}
\label{tab:map-recovery}
\end{table*}

\begin{table*}[p]
\centering
\footnotesize
\setlength{\tabcolsep}{3.5pt}
\begin{tabularx}{\textwidth}{@{}p{4.6cm}lllX@{}}
\toprule
\textbf{Work} & \textbf{A} & \textbf{Cond.} & \textbf{Instr.} & \textbf{What is scored} \\
\midrule
Audio-judge reliability \citep{sayyad2026lalmjudges} & C & AV & H, J & Per-dimension agreement and defect sensitivity of Gemini audio judges versus three human raters \\
CDS \citep{bokkahallisatish2026cds} & C, T & RS, LS & H & Distributional distance of conversational features to human references, validated against listener preferences \\
Distribution-based turn-shift evaluation \citep{kubo2026binaryturn} & T & NE & H & Wasserstein distance between label-conditioned predicted turn-shift probability distributions and reference distributions \\
Judge study \citep{purwar2026judges} & C, R, G & LS & H, J & Agreement of model judges with human raters on voice-agent conversations \\
LALM-as-a-Judge \citep{ivry2026lalmasajudge} & C & LS & H, J & Judge sensitivity, severity-order specificity and turn-position bias on unsafe spoken dialogues \\
Prosody references \citep{hallur2026prosodyref} & C & RS & -- & Percentile deviations and out-of-regime flags for prosody and pausing metrics \\
Speech-judge shortcuts \citep{huo2026speechjudgeaudit} & C & RS, AV & H, J & Whether speech judges reward irrelevant acoustic cues and whether rationales name them \\
TurnNat \citep{zhang2026turnnat} & T & RS & H & Metric discrimination of natural versus timing-perturbed dyadic dialogue clips \\
User-behavior indicators \citep{inoue2024userbehaviors} & T & NE, RS, LS & H & Relationship between observable user behaviours and subjective evaluation scores per task \\
\bottomrule
\end{tabularx}
\caption{Meta-evaluation: the 9 works whose primary target is meta-evaluation. Codes as in \cref{app:codebook}; ``--'' means none reported.}
\label{tab:map-meta}
\end{table*}

\begin{table*}[p]
\centering
\footnotesize
\setlength{\tabcolsep}{3.5pt}
\begin{tabularx}{\textwidth}{@{}p{4.6cm}lllX@{}}
\toprule
\textbf{Work} & \textbf{A} & \textbf{Cond.} & \textbf{Instr.} & \textbf{What is scored} \\
\midrule
APEX-Voice \citep{mathur2026apexvoice} & R, G & LS & H, J, S, Sim & End-to-end workflow success (terminal state, process, actions, artifact) and artifact field accuracy \\
Audio2Tool \citep{pahwa2026audio2tool} & R, G, M & AV, LS & -- & Tool-call correctness under compositional and acoustic complexity \\
DuplexWorld \citep{bhosale2026duplexworld} & C, T, R, G & AV, LS & J, S, Sim & Goal-state and gold-action task success, plus turn-taking and speech naturalness \\
EVA-Bench \citep{bogavelli2026evabench} & C, T, R, G & AV, LS & H, J, S, Sim & Separate accuracy and experience composites over repeated simulated trials \\
EVI enrolment/verification/identification tasks \citep{spithourakis2022evi} & G & NE, RS & S & Precision/recall/F1 for enrolment profile-field extraction; accuracy of accept/reject verification decisions and of identification against the KB \\
FDB-v3 \citep{lin2026fdbv3} & T, R, G & RS & J, S & Tool use from real disfluent speech; judged response quality \\
MTVA-Bench \citep{mishra2026mtva} & C, R, G & NE, LS & J, S, Sim & Language model inside a noisy cascade; hidden backend state and actions \\
SDN (Situated Dialogue Navigation) benchmark \citep{ma2022dorothie} & R, G & RS & S, Sim & Move accuracy and dialogue slot F1-score for UfN/RfN tasks; action accuracy (trajectory deviating less than 15 degrees counts as accurate) \\
SpeechGym \citep{speechgym2026} & R, G & LS & S, Sim & Audio-native agentic tasks under an established benchmark's success check \\
Talk2Agent \citep{chiba2026talk2agent} & G & -- & J, S & Downstream agent execution success and grader-weighted retention of task intentions \\
$\tau$-Elicitation \citep{ray2026tauelicitation} & R, G & AV, LS & H, J, S, Sim & Exact match of final database value to gold, robust across three environments \\
$\tau$-Multilingual \citep{tau2026multilingual} & C, T, G, M & NE, LS & H, J, S, Sim & $\tau$-Voice tasks in five further languages, with native-speaker review \\
$\tau$-Voice \citep{ray2026tauvoice} & T, R, G, M & AV, LS & H, J, S, Sim & Full-duplex task completion verified against database state \\
Text-to-voice tool calling \citep{laskar2026texttovoice} & G & AV & H, J & Tool-call correctness (AST soft accuracy, When2Call F1) on TTS audio versus text \\
VAmoS \citep{meyer2026vamos} & T, G & LS & J, S, Sim & Assertions on the interaction trace and seeded database state \\
VoiceAgentBench \citep{jain2025voiceagentbench} & C, T, G & NE, AV, LS & H, J & Tool selection, arguments, workflows, and safety in English and Indic languages \\
\bottomrule
\end{tabularx}
\caption{Grounded outcome: the 16 works whose primary target is grounded outcome. Codes as in \cref{app:codebook}; ``--'' means none reported.}
\label{tab:map-grounded}
\end{table*}

\begin{table*}[p]
\centering
\footnotesize
\setlength{\tabcolsep}{3.5pt}
\begin{tabularx}{\textwidth}{@{}p{4.6cm}lllX@{}}
\toprule
\textbf{Work} & \textbf{A} & \textbf{Cond.} & \textbf{Instr.} & \textbf{What is scored} \\
\midrule
Addressee recognition \citep{inoue2025addressee} & M & NE & H & Accuracy of predicting the addressed participant or no addressee \\
Duplex-MPE \citep{duplex2026mpe} & C, T, R, M & LS & H, J & Response initiation, answer accuracy, silence preservation, and stopping \\
HEAR (speaker) \citep{lee2026hear} & C, M & RS, AV & H & Speaker-attributed reasoning with counterfactual voices \\
M3-SLU \citep{kwon2025m3slu} & M & RS, LS & J & Speaker-attributed QA correctness and same-speaker T/F accuracy \\
Meeting turn-taking \citep{fukuda2026meetingturntaking} & T, M & RS & H & Accuracy and F1 for addressee, turn-change and next-speaker prediction \\
MP-Bench \citep{shih2026mpbench} & C, T, M & AV & H, J, Sim & Multiparty turn-taking and response appropriateness \\
MSI-Bench \citep{xiong2026msibench} & C, T, R, G, M & NE, AV, LS & H, J & Speaker-scoped memory, instructions, and expected tool calls \\
MSU-Bench \citep{sun2026msubench} & M & NE & H & Multiple-choice accuracy on speaker grounding and multi-speaker dialogue reasoning \\
Next speaker prediction (Teidan) \citep{mori2026nextspeaker} & T, M & NE, LS & H & Macro-F1 over addressed speakers, interruption avoidance rate and speaking opportunity detection rate \\
SocialOmni \citep{xie2026socialomni} & C, T, M & -- & H, J & Speaker attribution, turn-entry classification and judged continuation quality with coverage \\
TPI-Bench \citep{lee2026tpibench} & R, M & RS, AV & H, J & Interruption-response strategy following, helpfulness, and single- versus two-speaker discrimination \\
VGBench \citep{zhang2026vgbench} & G, M & RS, AV & J & Choice of silence, tool call, or spoken answer under side-talk and speaker switches \\
WearVox \citep{lin2026wearvox} & C, G, M & NE, RS, AV, LS & H, J & Turn accuracy on QA, tool calls and side-talk rejection; session translation score \\
\bottomrule
\end{tabularx}
\caption{Multiparty conduct: the 13 works whose primary target is multiparty conduct. Codes as in \cref{app:codebook}; ``--'' means none reported.}
\label{tab:map-multiparty}
\end{table*}

\begin{table*}[p]
\centering
\footnotesize
\setlength{\tabcolsep}{3.5pt}
\begin{tabularx}{\textwidth}{@{}p{4.6cm}lllX@{}}
\toprule
\textbf{Work} & \textbf{A} & \textbf{Cond.} & \textbf{Instr.} & \textbf{What is scored} \\
\midrule
Audio MultiChallenge \citep{gosai2025audiomultichallenge} & C, R & RS, LS & H, J & Rubric pass rate on memory, instruction retention, self coherence and voice editing \\
Back-transcription NLU-robustness method \citep{kubis2023backtranscription} & C & -- & -- & Comparison of NLU performance degradation and error-category impact between synthesized-speech back-transcription and recorded-audio back-transcription \\
C3 \citep{ma2025c3} & C & NE, LS & H, J & Judge-assessed accuracy on ambiguity, omission, coreference and multi-turn items \\
ContextDialog \citep{kim2025contextdialog} & C & LS & H, J & GPT-4o-mini 5-point recall score for answers about past user and model utterances, with and without RAG \\
Dialog Quality Annotation (DQA) model \citep{komma2023dqa} & C & RS & H & Agreement of the dialog-level quality prediction model with held-out expert annotations, compared to a turn-aggregation baseline and to end-user satisfaction labels \\
DSB-DG \citep{mathur2026dsbdg} & C, T & LS & J & Factual support by supplied documents, hallucination rate, multi-turn retention, first-token latency \\
DSTC10 spoken dialogue state tracking / knowledge-grounded dialogue benchmark \citep{kim2021howrobust} & C & RS & -- & Dialogue state tracking joint goal accuracy and knowledge-grounded response quality, written-trained models evaluated on spoken test data, with and without n-best ASR hypothesis fusion \\
HEAR (bias) \citep{yan2026hearbias} & C & RS, AV & -- & Voice-conditioned demographic bias over 87k human recordings \\
MMedFD \citep{chen2025mmedfd} & C & NE, RS, LS & H, J & Recognition in healthcare full-duplex sessions; judged responses \\
MTalk-Bench \citep{du2025mtalkbench} & C, M & RS, AV, LS & H, J & Semantic, paralinguistic and ambient-sound handling via arena Elo and binary rubrics \\
Next-utterance DA classification (MapTask) \citep{miyazawa2026paranuc} & C & NE, RS & H & MRR@5 and top-1 accuracy of next dialogue-act prediction under paralinguistic-agnostic, perturbed and aware conditions \\
RealTalk-CN \citep{wang2026realtalkcn} & C & NE, RS, AV, LS & J & Intent accuracy, slot F1/JGA and judged chat quality, split by disfluency, domain, speaker age/region and modality switching \\
RoleBreak \citep{wang2026rolebreak} & C & AV, LS & H, J & Criterion pass rates, persona/safety first-failure turn, naturalness, vocal emotion match \\
SCB \citep{vesessook2026scb} & C & LS & -- & Final-answer accuracy under full, concatenated and incrementally sharded spoken problems \\
SOVA-Bench \citep{hou2025sovabench} & C & -- & J & Knowledge, recognition, spoken QA, paralinguistics, and semantic/acoustic quality of generated speech \\
SPEARBench \citep{thebaud2026spearbench} & C, T & RS & J & Answer latency, interruptions, speech quality, dialect, emotion and stance versus human answers \\
SpeechJBB \citep{ceccatelli2026speechjbb} & C & NE & H, J & Spoken jailbreak safety under code switching \\
TalkArena \citep{li2025mindthegap} & C & RS & H & User pairwise preferences between speech-in text-out model responses; static-benchmark correlation \\
TELEVAL \citep{li2025televal} & C & NE, RS, AV, LS & H, J & Semantic accuracy under acoustic variation and cue-appropriate response quality \\
URO-Bench \citep{yan2025urobench} & C, T & NE, LS & J & Task accomplishment, speech quality, ASR-WER and first-packet latency across S2S test sets \\
VCB Bench \citep{hu2025vcb} & C & NE, RS, AV, LS & H, J & Instruction following, knowledge, and robustness on real Chinese speech \\
VocalBench \citep{liu2025vocalbench} & C, T & NE, AV, LS & H, J & Accuracy, LLM 1-5 scores, empathy, robustness preserve rate, latency across 14 capabilities \\
VoiceBench \citep{chen2024voicebench} & C & RS, AV & H, J & Response quality under speaker, environment, and content variation \\
Vox-Infinity \citep{cheng2026voxinfinity} & C & RS, LS & H, J & QA accuracy by distance of answer-supporting evidence in audio history \\
WildSpeech-Bench \citep{zhang2025wildspeech} & C & RS, AV & H, J & GPT-4o-mini 1-10 scores of ASR-transcribed speech responses, checklist-guided \\
\bottomrule
\end{tabularx}
\caption{Component quality: the 25 works whose primary target is component quality. Codes as in \cref{app:codebook}; ``--'' means none reported.}
\label{tab:map-component}
\end{table*}

\end{document}